\documentclass[review,12pt,3p]{elsarticle}

\usepackage{amsmath}
\usepackage{amssymb}
\usepackage{verbatim}  
\usepackage{siunitx}
\usepackage{algorithmic}
\usepackage{algorithm}
\usepackage{array}
\usepackage{eqparbox}

\begin{document}

\begin{frontmatter}

\title{Deep Reinforcement Learning: An Overview}

\author[label1]{Seyed Sajad Mousavi}
\address[label1]{The College of Engineering and Informatics\\
National University of Ireland, Galway, Republic of Ireland}
\ead{s.mousavi1@nuigalway.ie}

\author[label1]{Michael Schukat}
\ead{michael.schukat@nuigalway.ie}

\author[label1]{Enda Howley}
\ead{schukat,ehowley@nuigalway.ie}

\begin{abstract}
In recent years, a specific machine learning method called deep learning has gained huge attraction, as it has obtained astonishing results in
broad applications such as pattern recognition, speech recognition, computer vision, and natural language processing. Recent research has also been shown
that deep learning techniques can be combined with reinforcement learning methods to learn useful representations for the problems with high dimensional
raw data input. This paper reviews the recent advances in deep reinforcement learning with a focus on the most used deep architectures such as autoencoders,
convolutional neural networks and recurrent neural networks which have successfully been come together with the reinforcement learning framework. 
\end{abstract}

\begin{keyword}
Reinforcement learning \sep Deep Reinforcement learning \sep Deep leaning \sep Neural networks \sep MDPs \sep Observable MDPs 
\end{keyword}

\end{frontmatter}


\section{Introduction}

Reinforcement Learning (RL) \cite{kaelbling1996reinforcement,sutton1998introduction} is a branch of machine learning in which an agent learns from interacting with an environment. Before an agent or robot (software or hardware) can select an action, it must have a good representation of its environment \cite{kober2013reinforcement}. Thus, perception is one of the key problems that must be solved before the agent can decide to select an optimal action to take. Representation of the environment might be given or might be acquired. In reinforcement learning tasks, usually a human expert provides features of the environment based on his knowledge of the task. However, for some real world applications this work should be done automatically, since automatic feature extraction will provide much more accurate. There are several solutions to deal with this challenge, such as Mont Carlo Tree search \cite{vien2013monte}, Hierarchical Reinforcement Learning \cite{mousavi2014automatic} and function approximation \cite{sutton1998introduction}. Hence, we are interested in the ways that can be used to represent the state of the environment.

Deep learning attempts to model high-level abstractions in data using deep networks of supervised and/or unsupervised learning algorithms, in order to learn from multiple levels of abstractions. It learns hierarchical representations in deep architectures for classification.

In recent years, deep learning has gained huge attraction not only in academic communities (such as pattern recognition, speech recognition, computer vision and natural language processing), but it has also been used successfully in industry products by tech giants like Google (Google’s translator and Image search engine), Apple (Apple's Siri), Microsoft (Bing voice search) and other companies such as Facebook and IBM. Recent research has also shown that deep learning techniques can be used to learn useful representations for reinforcement learning problems \cite{mattner2012learn,mnih2013playing,levine2016end,bohmer2015autonomous,mnih2015human}. Combining the RL and deep learning techniques enables RL agent to have a good perception of its environment.
 
The aim of this study is to outline and critically review all significant research done to date in the context of combining reinforcement learning algorithms and deep learning methods. The research will review both supervised and unsupervised deep models that have been combined with RL methods for environments which might be partially observable MDPs or not. This study will also present recent outstanding success stories of the combined RL and deep learning paradigms, which led to the introduction of a novel research route called deep reinforcement learning, to overcome the challenges in learning control policies from high-dimensional raw input data in complex RL environment.

The rest of this paper is organized as follows. Sections \ref{sec:reinlearning} and \ref{sec:deeplearning} give a brief review of reinforcement learning and deep learning (focused on three commonly used deep learning architectures with reinforcement learning framework), respectively. Section \ref{sec:deepsupunsup} presents the state of the art techniques that combine reinforcement learning and deep learning with a rough categorization of the combination of deep supervised learning models with RL and the combination of deep unsupervised learning models with RL. Finally, Section \ref{sec:conclusion} summarizes the paper.

\section{Reinforcement Learning}
\label{sec:reinlearning}
Reinforcement Learning \cite{kaelbling1996reinforcement,sutton1998introduction} is a branch of machine learning in which an agent learns from interacting with an environment.An RL framework allows an agent to learn from trial and error. The RL agent receives a reward by acting in the environment and its goal is learning to select the actions that maximize the expected cumulative reward over time. In other words, the agent, by observing the results of those actions that it is taking in the environment, tries to learn an optimal sequence of actions to execute in order to reach its goal.

A reinforcement learning agent can be modelled as a Markov decision process (MDP). If the states and action spaces are finite, then the problem is called a finite MDP. Finite MDPs are very important for RL problems and much of literatures have assumed the environment is a finite MDP in their works.

The way an RL agent acts in the finite MDP framework as follows: The learning agent interacts with the environment by executing actions and receiving observations and rewards. At each time step $t$, which ranges over a set of discrete time intervals, the agent select an action $a$ from a set of legal actions $A={1,2,\ldots,k}$ at state $s_t\in S$, where $S$ is the set of possible states. Action selection is based on a policy. The policy is a description of the behaviour of the agent which tells the agent which actions should be selected for each possible state. As a result of each action, the agent receives a scalar reward $r_t \in \mathbb{R}$, and observes next state $s_{t+1} \in S$ at one step time later. The probability of each possible next state $s_{t+1}$ comes from a transition distribution which is $P(s_{t+1}|s_t,a_t); s_{t+1},s_t \in S, a_t \in A(s_t)$. Similarity, the probability of each possible reward $r_t$ comes from a reward distribution $P(r_t|s_t,a_t); s_t \in S, a_t \in A(s_t)$. Hence, the expected scaler reward received, $r_t$ , by executing action $a$ in current state $s$ is calculated based on $E_{P(r_t|s_t,a_t)}(r_t|s_t=s,a_t=a)$.

The aim of the learning agent is to learn an optimal policy $\pi$, which defines the probability of selecting action $a$ in state $s$, so that with following the policy the sum of the discounted rewards over time is maximized. The expected discounted return $R$ at time $t$ is defined as follows:
\begin{equation} \label{eq:eq1} R_t = E[r_t +  \gamma r_{t+1} + \gamma^2 r_{t+2} + \ldots ]=E[\sum_{k=0}^{\infty}\gamma^k r_{t+k}], \end{equation}

Where $E[.]$ expectation with respect to the reward distribution and $0<\gamma<1$ is called the discount factor. With regard to the transition probabilities and the expected discounted immediate rewards, which are the essential elements for specifying dynamics of a finite MDP, action-value function $Q^\pi (s,a)$ is defined as follows:

\begin{equation} \label{eq2} Q^\pi (s,a) = E_\pi [R_t|s_t= s, a_t = a] = E_\pi [\sum_{k=0}^{\infty}\gamma^k r_{t+k}|s_t=s, a_t = a], \end{equation}

The action-value function $A_\pi (s,a)$ for an agent is the expected return achievable by starting from state $s$, $s \in S$, and performing action $a$, $a \in A$and then following policy $\pi$, where $\pi$ is a mapping from states to actions or distributions over actions.

With unfolding the equation \ref{eq2} it is clear that it satisfies a recursive property, so that the following iterative update can be used for the estimation of action-value function:

\begin{equation} \label{eq3}  \begin{split}
{Q^\pi}_{i+1} (s,a) = E_\pi[r_t + \gamma \sum_{k=0}^{\infty} \gamma^k r_{t+k+1}|s_t=s,a_t= a] \\
                = E_\pi [r_t + \gamma {Q^\pi}_i (s_{t+1}=s',a_{t+1}= a')|s_t=s, a_t=a], \end{split} \end{equation}

For all $s, s' \in S$ and $a, a' \in A$, in Eq. \ref{eq3}, both states a relationship between the value of an action in a state and the values of its next actions which can be performed It also cites the way of estimating the value based on its subsequent ones.

Reinforcement learning agent wants to find a policy which achieves the greatest future reward in the course of its execution. Hence, it must learn an optimal policy $\pi^*$, a policy which is resulted to an expected value greater than or equal of following other policies for all states, and as a result, an optimal state-value function $Q^*(s,a)$. In particular, an iterative update for estimating the optimal state-value function is defined as follows:

\begin{equation} \label{eq:eq4} Q_{i+1}(s,a) = E_\pi[r_t +  \gamma max_{a'}Q_i(s',a') |s,a], \end{equation}

Where it is implicit that $s, s' \in S$ and $a, a' \in A$. The iteration converges to the optimal action-value function, $Q^*$ as $i \rightarrow \infty$ and called value iteration algorithm \cite{sutton1998introduction}.

In many real problems, number of states and actions are very large and use of classical solution (state-action table to store the values of state-action pairs) is
impractical. On way to deal with is use of function approximator as estimator of action-value function. The approximate value function is parameterized $Q(s,a;\theta)$ with
parameter vector $\theta$. Usually gradient-descent methods are used to learn parameters by trying to minimize the following loss function of mean-squared error in Q-values:

\begin{equation} \label{eq:eq5} L(\theta) = E_\pi[(r + \gamma max_{a'}Q(s',a';\theta)-Q(s,a;\theta))^2], \end{equation}

Where $r + \gamma max_{a'}Q(s',a';\theta)$ is the target value. Differentiation of the loss function with respect to its parameters $\theta$ lead to the following gradient:

\begin{equation} \label{eq:eq6} \frac{\partial L(\theta)}{\partial \theta} = E[(r + \gamma max_{a'}Q(s',a';\theta)-Q(s,a;\theta)) \frac{\partial Q(s,a;\theta)}{\partial \theta}, \end{equation}

This is the way the gradient-based methods are applied. Typically, the gradient above is optimised by stochastic gradient descent method. The approximate function can
be a linear function or a non-linear function (for example a neural network) of the parameters $\theta$. Until recently, the majority of work in reinforcement learning utilized
linear function approximatiors because the convergence guarantees that they provide. More recently, by alleviating the convergence problems, not only typical neural networks,
for example multi-layer perceptrons (MLP), have been common to use as function approximators for large reinforcement learning tasks, but deep neural networks such as convolutional neural networks and recurrent neural networks is used \cite{riedmiller2005neural,oh2015action,mnih2013playing}.

\section{Deep Learning}
\label{sec:deeplearning}
Obtaining a good performance of a machine learning technique is highly dependent on having good representation of input data. Hence, the pre-processing of the data (i.e.,
feature learning) is a critical step in the process of creating the machines which are learned through the observation of the data. The feature engineering process is a way to
take advantage of the knowledge of the domain experts in order to extract hand-crafted features and to reduce dimension of features of the input data. Effectiveness of the
shallow learning models such as support vector machines (SVMs) and logistic regression are dependent on feature learning. This process is important but very time
consuming and difficult to do. It would be better to have algorithms that facilitate the problem. Deep learning techniques are one the best solutions to deal with high
dimension data and extract discriminative information from the data. Deep learning algorithms have the capability of automating feature extraction (the extraction of
representations) from the data. The representation are leant through the data are fed directly into deep nets without human knowledge (i.e., automated feature extraction).
This key aspect of deep learning architectures is led to progress towards those algorithms that are the goal Artificial Intelligence (AI), understating the world around
independent of expert knowledge and interference \cite{bengio2013representation}.

In summary, deep learning attempts to model high-level abstractions in data using deep networks of supervised and/or unsupervised learning algorithms, in order to learn
from multiple levels of abstractions. It learns hierarchical representations in deep architectures for different tasks such as classification.

Deep learning models contain multiple layers of representations. Indeed, it is a stack of building blocks such as autoencoders, Restricted Boltzmann
Machines (RBMs) and convolutional layers. During training, the raw data is fed into a network consisting of multiple layers. The output of the each layer which is nonlinear
feature transformations, is used as inputs to the next layers of the deep network. The output representation of the final layer can be used for constricting classifiers or those applications which can have the better efficiency and performance with abstract representation of the data in a hierarchical manner as inputs. Each layer by applying a nonlinear transformation on its input try to learn and extract underlying explanatory factors. Consequently, this process is led to learn a hierarchy of abstract representations. For example, using deep leaning algorithm in image processing applications, the first layer is provided by the image pixels which can lead to learn the edges of different object
in image. The second layer uses the representations provided by the first layer to lean complex features such as object parts (combination of edges). The third layer composes
object parts (more complex features) to figure out object models. The example shows the hierarchical learning power of the abstracted representations by a deep learning algorithm are able to recognize objects in the image. For this reason, the deep learning approach can be considered as a kind of representation learning algorithms \cite{bengio2013representation}.

The use of deep neural networks allow the deep learning methods to be very powerful tools in solving real problem. However, learning the parameters (deep nets with many
hidden layers are led to have millions of parameters to learn) in a deep architecture is a difficult optimization task which imposes very high computational complexity \cite{lecun1998gradient}. Fortunately, with emergent of advanced parallel processing technologies like GPU this problem has alleviated somewhat. 

In the following sections, we will introduce a brief overview of three kind of deep neural networks which recently have successfully been had the most use in combination with RL. These deep architectures are included autoencoders (stacked [denoising] autoencoders), convolutional neural networks (CNNs) and recurrent neural networks (RNNs).

\subsection{Autoencoder}

The deep autoencoders are one of the most notable works in unsupervised deep feature learning. They are a type of artificial neural networks which try to learn a representation
of original data. An autoencoder, autoassociator or Diabolo network \cite{bengio2009learning} is trained to learn effective encoding of input date so as to the input can be reconstructed from the encoding. In fact, the target output of the network is the same of the input. It typically has three layer: input layer, used for input feature vectors; hidden layer, used for representing mapped features, and output layer which use to represent reconstructed input. Like other neural networks, the backpropagation methods are used to learn parameters, typically the stochastic gradient descent method. If the architecture of the autoencoder only contains one linear hidden layer and the mean squared error criterion, as the loss function, is used to train the network, autoencoder will performs like principal components analysis (PCA) method and will learn the first k principle components of the data \cite{bengio2009learning}. In order to get more benefits of the autoencoder rather than dimensionality reduction method, non-linear hidden units are used in the hidden layer.

To achieve more expressive power of autoencoders, stacked auto-encoders are constructed by stacking up autoencoders in tandem \cite{bengio2007greedy}. The output of each autoencoder is connected to the input of the next autoencoder. There is also a variant of autoencoder called demonising autoencoder \cite{vincent2008extracting}. The denoising autoencoder minimizes the reconstruction error of corrupted versions of the input (random noises are added to input data) and tries to recover the original input data, i.e., without distortions. There are two key ideas behind of this approach. First, the use of denoising will lead to a higher level representations which are more robust and stable when there is the noises in input data. Second, the use of the denoising task will force to extract those features which have more effect on useful structure of the input distribution \cite{vincent2010stacked}.

\subsection{Convolutional Neural Networks}

Convolutional Neural Networks (CNNs) are categorised in the class of supervised deep feature learning models. Perhaps one of the first research on CNNs is work done by \citet{lecun1989backpropagation}. They used CNNs to recognise handwritten characters. By advances in power of computing devises they were able to apply CNNs on other applications such as object recognition and detection in image and speech recognition and time series \cite{lecun1995convolutional}.

Convolutional networks are constructed of many layers and connections aim to learn hierarchical feature representation. To be invariant on some degree of translational and distortional of input feature vectors, three main strategies are applied, local receptive fields, shared weights and spatial or temporal subsampling \cite{lecun1998gradient}.

\begin{figure}
  \includegraphics[width=\linewidth]{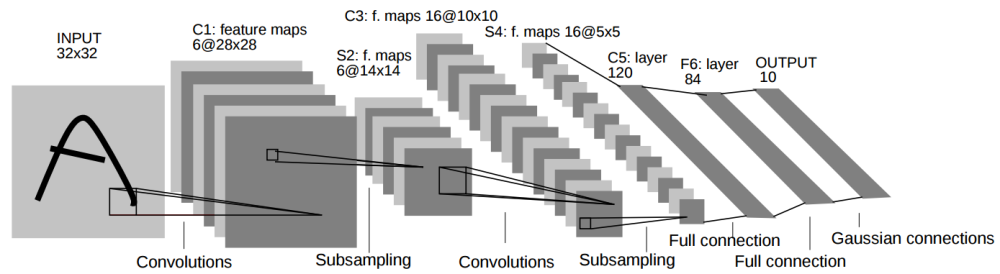}
  \caption{Architecture of a typical convolutional neural network which illustrates how the
convolution and the subsampling layers are connected together and are fallowed by fully
connected layers to construct a CNN, adapted from \cite{lecun1998gradient}.}
  \label{fig:lenet5}
\end{figure}

Figure \ref{fig:lenet5} shows a basic architecture of the CNNs. The first two layers of the CNNs are convolutional and subsampling layers. Convolutional layer performs convolution operation to create feature maps. It uses local receptive fields (small filter sizes) and shared weights (filter maps with equal size,) to bring distortion invariance attribute. The output of the convolution operation is then passed through a nonlinear activation function which is followed by a subsampling layer. Subsampling layer performs local averaging (or max-pooling \cite{scherer2010evaluation}) that reduces dimensionality of proceeding feature maps while keeping distortion invariance. Sequence of convolutional and subsampling layers can be used in tandem in the architectures of CNNs of a specific task. The output of the final subsampling layer is fed to some fully connected layer for classification or recognition tasks. An interested reader is referred to \cite{lecun1998gradient} paper for more details on CNNs.

However, in order to CNNs to be applicable in real applications, especially in applications with high dimensional input data such as image and speech processing,
and achieve reasonable performance in comparison with shallow learner methods, this kind of deep networks must be provided with the large amount of data. In addition,
with so many parameters to train of the deep architecture, the high performance computing power is also required. Until recently they had not been widely used
because of the mentioned requirements. These problems can now be dealt easily with emerging highly parallel Graphical Processing Units (GPU) and becoming available
large data sets. For instance, in the area of image and vision research, with presenting of ImageNet database \cite{deng2009imagenet}, a large-scale image database which consists of millions labelled high resolution images in thousands of categories, and parallel GPUs, deep CNN architectures could outperform state of the art algorithms in pattern recognition.

\subsection{Recurrent Neural Networks}

Another deep supervised feature learning (as well as unsupervised) algorithm which is used for sequential information, where input data are depended to each other in the way
they are coming out (data stream) or have located (words in a sentence), is recurrent neural networks (RNNs). Unlike feedforward neural networks (FNNs), recurrent neural
networks (RNNs) have feedback connections which allows them to have internal states. This means that they have a memory which can keep information about previous
inputs, enabling them to be useful for those applications such as speech recognition which has temporal and sequential data \cite{deng2013new}.

To learn long-term dependencies in sequential data using vanilla RNN architecture gives rise gradient vanishing (the magnitude of the error gradients vanish exponentially
during training, which makes it impossible for the RNN to learn correlation between temporally long-term events) or gradient explosion (a large increase in the magnitude
of the gradients during training, where long-term components exponentially grow and dominate the gradients of short term ones) problems \cite{bengio1994learning}. Other types of RNNs such as long short term memory (LSTMs) \cite{hochreiter1997long} was introduced to deal with these problems. LSTMs are able to learn very long-term dependencies. They could outperform alternative RNNs and Hidden Markov Model (HMM) approaches which were state of the art in sequence learning \cite{deng2013new}.

\section{Deep Supervised and Unsupervised Learning Models for Reinforcement Learning}
\label{sec:deepsupunsup}
The most well-known reinforcement learning algorithm which uses neural networks (but no deep nets, i.e., there is only one hidden layer) is the world-class RL
backgammon player named TD-Gammon, which gained a score equal to human champions by playing against itself \cite{tesauro1995td}. TD-Gammon uses TD (lambda) algorithm
\cite{sutton1998introduction} to train a shallow neural net to learn to play the game of backgammon. However, later attempts to use the same method for other games such as chess, Go and checkers were not successful.

With riving interest in research works on deep learning in the middle of the 2000s decade, the promise to use neural networks as function approximator both for the state
value function $V(s)$ and the action-value function $Q(s, a)$ in visual based RL tasks came back. In the following sections we introduce those works that have used the deep neural nets in combination with reinforcement learning framework to improve the performance of learning control policies with emphasizing on those that are fed with raw
input data.

\subsection{Combination of RL Techniques with Supervised Learning Approaches of Deep Neural Networks}

The work in \cite{riedmiller2005neural} has proposed a model free method called Neural Fitted Q learning (NFQ). NFQ update the weights of a multilayer perceptron by RPROP algorithm \cite{riedmiller1993direct},a batch learning method for training the neural networks which is very fast in comparison with other supervised leaning methods, for regressing the value function where updating is accomplished offline. The update is based on an entire set of transition experiences which has triple form $(s, a, s’)$ where s is the current state, a is the selected action and s’ is the next state, which is the result of taking action a. Since updating is performed offline, the transition experiences are collected before. Indeed, they are acquired by interacting with a real or simulated environment. The proposed method has two steps (1) collecting learning set and (2) doing a batch update for training a multilayer perceptron by RPROP algorithm. Since it use a batch update the computational complexity of doing this update in each iteration is corresponding with the size of training set \cite{mnih2013playing}.

Since the most of the resent approaches that combine deep learning and RL use challenging environments introduced the Arcade Learning Environment (ALE) \cite{bellemare2013arcade}, we first introduce ALE. ALE provides an environment that emulate the Atari 2600 Games. Atari 2600 presents a very challenging environment for reinforcement learning that has a high dimensional visual input ($210 \times 160$ RGB video at 60 Hz) which is a partial observable observation. It provides a range of interesting games that the proposed methods can be tested, where the agent use the methods for playing the games.

More recently, researchers in DeepMind technologies have developed an approach called Deep Q learning Network (DQN) \cite{mnih2013playing} which benefits from advantages of deep learning for abstract representation in learning optimal policy, i.e. selecting actions in such a way that maximize the expected value of the cumulative sum of rewards. It is an extension of the previous work Neural Fitted Q-Learning (NFQ) \cite{riedmiller2005neural}. DQN combine a deep convolutional neural network with the simplest reinforcement learning method (Q-learning) to play several Atari 2600 computer games only by watching the screen.

Combing model-free reinforcement learning algorithms such as mere Q-learning algorithm with neural networks causes some stability issues and makes to be diverged.
There are two main reasons for these issues, e.g., (1) subsequent states in RL tasks are much correlated, (2) the policy is changing frequently, it is because of slight changes in Q-values. DQN for dealing with these issues provides some solutions. For the correlated states issue, it utilizes the approach introduced in \cite{lin1993reinforcement} named experience replay. In the process of learning, DQN store agent's experience $(s_t,a_t,r_t,r_{t+1})$ at each time step into a date set $D$, where $s_t$, $a_t$ and $r_t$, respectively the state, selected action and received reward at time step $t$ and $s_{t+1}$ is state at the next time step. For updating Q-values, it uses stochastic minibatch updates with uniformly random sampling from experience replay memory (previous transitions) at training time. This work break strong correlations between consecutive samples, and for instability in the policy, the network is trained with a target Q-network to obtain consistent Q-learning targets by fixing weight parameters used in Q-learning target and updating them periodically.

Until recently the proposed method achieved the best real time agents. In some games its strategy outperformed the human player and achieved state of the art performance
on many Atari games with the same network architecture or hyperparameters. Several factors have been involved for getting the significant results, while the previous
works had not considered \cite{bohmer2015autonomous}. First, advances in computing power, especially highly paralleled Graphical Processing Units (GPU) technology which has enabled training the deep neural networks with thousands of weight parameters. Second, DQN has used a large deep CNN which it has been made better representation learning. Third, DQN has used experience replay for the correlated states problem.

However, using deep neural networks need sufficient data to be fed into network to learn better representations and as a result getting good performance. Hence, applying
this approach in real environment such as robotics is very challenging and difficult since performing a large number of episodes to collect samples is source consuming
and even not possible.

The work done by \citet{guo2014deep} has shown better results in comparison with DQN's performance. It uses the offline Mont Carlo tree search planning to provide
training data for a convolutional neural network. Indeed, they have developed some methods which benefit from deep learning nets for abstract representation and
model-free RL by utilizing UCT-based planning method \cite{kocsis2006bandit} to generate input data for the CNN.

Like DQN this work also uses ALE framework as testbed for the proposed methods. It outperforms DQN in several games of Atari 2600. In order to achieve these
results UCT needs significant time between actions \cite{guo2014deep}. In addition, planning-based approaches are slow for real time play.

The goal of the research in \cite{levine2016end} is training the perception and control systems jointly rather than each phase is trained separately, in order to get better
performance. To reach this purpose, learning a policy which do both the perception and the control jointly, they utilized deep convolutional neural networks. The CNNs get
raw images from a PR2 robot’s camera and output the policy. The policy is a conditional distribution of a Gaussian which determines a probability distribution over
actions with regard to given the observation of environment. The authors evaluated their method with comparing to other policy research approaches on various tasks such
as hanging a coat hanger on a clothes rack, stacking Lego blocks on a fixed base, screwing caps onto pill bottles, etc. and has shown significant results (for more
information see \cite{levine2016end}). 

Developing an artificial agent which can play in variety of games is still a big challenge in AI. One type of these challenging games is board games such as the
two-player game of Go in which the goal is to surround more territory than the opponent. In \cite{gruttner2010multi}, the authors have developed a method to play the Go game on small
boards with combining the benefits of two approaches multi-dimensional recurrent neural networks (MDRNNs) and long short-term memory (LSTMs). The proposed
method benefit from the feature of MDRNN where it can use provided information of the two space dimensions of the game board. Moreover, by integrating the LSTM in
MDRNN, the vanishing gradient problem for RNNs \cite{hochreiter2001gradient} has solved as well. To train the networks they used policy gradients with parameter-based exploration method \cite{sehnke2010parameter}, a model-free reinforcement learning for POMDPs problems which has outperformed Evolution Strategies \cite{beyer2002evolution}. Notably, \cite{clark2015training} as well as have used CNNs for playing Go game which input data was raw visual pixels. Their proposed methods have resulted state of the art performance to the problem of predicting the moves made by expert Go players. However, combining CNNs and RL framework to deal with Go game might lead to better improvement.

Similar to the past works in visual based RL domain, the research \cite{koutnik2013evolving} receives high dimensional visual inputs and learns optimal policies using end-to-end reinforcement learning. It utilized a compressed recurrent neural networks which uses evolutionary algorithms for evolving the neural network as the action-value function approximate. It successfully used to two challenging tasks such as the TORCS race car driving with high-dimensional visual data streams and the visual Octopus Arm task.

The problem of video prediction, is another domain that combing deep learning and RL approaches can be an optimal solution. One notable work is \cite{oh2015action}. It introduces two deep neural networks architectures which integrate convolutional neural networks, recurrent neural networks and RL in order to predict action-conditional frames, the next frames in video depend on preformed actions in previous time step, the work done by \citet{guo2014deep} used slow UCT to predict future frames. They have shown that using their architectures in some Atari games can extract the features both spatial and temporal and generate 100-step action-conditional future frames without suffering of being diverged.

\subsection{Combination of RL Techniques with Unsupervised Learning Approaches of Deep Neural Networks}

There have been several attempts to learn representation which use the unsupervised learning techniques of deep neural networks in combination with RL. In the following
section, we will address some unsupervised deep neural networks which are used in order to learn compact low-dimensional feature space of the RL task. Solving
visual-based reinforcement learning tasks is usually divided into two steps. The first is, mapping high-dimensional input date into a low- dimensional representation (which
here, our focus is using the unsupervised learning methods of deep architectures). The second is, applying an approximation technique to the learned compacted feature space
for approximating the Q-value function or the control policies.  

The studied research by \citet{lange2010deep} to handle high-dimensional visual state spaces problem in RL task, presented the Deep Fitted Q-iteration (DFQ) algorithm in
which unsupervised training of deep auto-encoder networks are integrated with RL methods. DFQ algorithm at the first stage uses a deep auto-encoder to learn a low
dimensional presentation of the input state (image) and then at the second stage, applies NFQ algorithm \cite{riedmiller2005neural}, a batch-mode supervised learning, to estimate the Q-value function. DFQ algorithm was successfully applied to some continuous grid-world tasks which have had visual input.

Deep Fitted Q-iteration algorithm has also been successfully used to learn the control policy for two control tasks, a pole balancing and a racing slot car, respectively
by \citet{mattner2012learn} and \citet{lange2012autonomous}. They have followed two steps, (1) raw visual input which were captured by a digital camera, is fed into an auto-encoder network in order to reducing and condensing the input state space, and (2) to estimate the value function, in former a kernel based function approximator \cite{ormoneit2002kernel} has been applied and in latter, the ClusterRL approach \cite{lange2010deep} has been utilized. However, any function approximation methods can be used for accomplishing this step.

\subsection{Deep RL for Partially Observable MDPs (POMDPs) Environments}

In most real world applications the Markov assumption is not feasible, since real states are only partially observable and using only the current states for decision making
might not led to reach the optimal strategy. Unlike Markov decision processes (MDPs), POMDPs assume the input states of the RL agent are not complete and cannot contain
all necessary information to select the optimal next action. One way which can facilitate this inconsistency is memorizing the history of the past observations, for this
purpose \cite{mnih2013playing} have stacked a history of the last 4 frames that the agent has recently seen in their experiments when they have used Atari 2600 games as the testbed.

Recurrent networks are common solutions when the arbitrary background of the past events in a system is in need. RNNs are used as function approximators where they
can provide condensed feature spaces of the past events which have been seen thus far. Indeed, this combination, POMDP RL and DL RNNs, allows the agent to memorize
important previous observations \cite{schmidhuber2015deep}. One the first use of recurrent networks with RL is the research done by \citet{bakker2003robot} in robotics domain which enabled the RL robot with the memory capability through a Long Short Term Memory (LSTM) recurrent neural network, a special kind of the RNNs that can learn long term dependencies of states already seen \cite{hochreiter1997long}.

The proposed method by \citet{hausknecht2015deep} has adjusted Deep Q-network so as to it can be used in those environments which the observations may be noisy and incomplete (e.g., POMDPs environments). For this purpose they have integrated a LSTM with a DQN.

To test their works they introduced the Flickering Pong POMDP which is a modified version of the game of Pong (one of the Atari 2600 games). The game screen
is considered fully obscured with probability p = 0.5 to bring the condition of the POMDP for Pong game. In Flickering Pong environment, they have shown that a
recurrent deep Q-network (DRQN) with a single frame as input, has better performance in comparison with DQN with 4 frames and 10 frames as inputs. They have also proved
that DRQN is cable to generalize its policies to MDP environment, where the states are completely observable. However, it had not significant superiority in other Atari games
as benchmark, when it has been compared to DQN with the history of frames \cite{hausknecht2015deep}.

\section{Conclusions and Future Work in Deep Reinforcement Learning}
\label{sec:conclusion}
Deep learning models with great power of automatically extracting complex data representations from high-dimensional input data could outperform other state of the art of traditional machine learning methods. A major challenge in reinforcement learning is to learn optimal control policies in problems with raw visual input. Hierarchical feature extraction and learning abstracted representations of deep architectures, not only made the deep learning become a valuable tool for classification, but it has made it to be a great solution for the mentioned challenge in RL tasks as well.

In this paper, the focus was the role of deep neural networks as a solution for dealing with high-dimensional data input issue in reinforcement learning problems. We have presented recent advances in combing reinforcement learning framework and deep leaning models for both deep supervised and unsupervised learning networks. In particular, the deep architectures that have been most used in combination with RL such as deep convolutional networks, deep autoencoders and deep recurrent networks. In addition, appropriate deep networks for the problems with partially observable MDPs (POMDPs)
environment, have been discussed.

Despite of the significant works done to data in combining RL and DL, research on deep reinforcement learning is at its first steps and there are still many unexplored aspects of this combination. Also, their challenges in real application such as robotics, are yet unsolved and need more exploration to be done. More work is necessary on investigating deep architectures both for end to end leaning, which performs a direct approach to learn non-linear control policies, and deep state representation, which does dimension reduction to present low dimensional representations then try to approximate Q-values. Especially, developing those mechanisms which make the end to end learning
can be practical in real world application, those which doing a large number of actions is impossible.

Furthermore, an open problem that has not yet been addressed is how deep architectures can help deep reinforcement learning models to transfer knowledge (transfer learning). Indeed, how to use learned features by the deep networks for different tasks, without changing the network architectures.



\bibliographystyle{elsarticle-harv}

\bibliography{sample}

\end{document}